\documentclass[10pt,twocolumn,letterpaper]{article}

\usepackage{ijcb}
\usepackage{times}
\usepackage{epsfig}
\usepackage{graphicx}
\usepackage{amsmath}
\usepackage{amssymb}

\ijcbfinalcopy 

\ifijcbfinal\fi

\begin{document}

\title{Weakly Supervised Detection of Baby Cry}

\author{Weijun Tan$^{1,2}$, Qi Yao$^{1}$, Jingfeng Liu$^{1}$\\
$^{1}$Deepcam Information Technologies, Shenzhen, China\\
$^{2}$LinkSprite Technologies, Longmont, CO 80503, USA\\
{\tt\small {weijun.tan@deepcam.com},{weijun.tan@linksprite.com}}\\
{\tt\small {qi.yao@deepcam.com}}\\
{\tt\small {jingfeng.liu@deepcam.com}}
\and
{}
}

\maketitle

\begin{abstract}
   Detection of baby cry is an important part of baby monitoring. Almost all existing methods use supervised SVM, CNN or their varieties. In this work, we propose to use weakly supervised anomaly detection to detect baby cry. In this weak supervision, we only need weak annotation if there is cry in an audio file. We design a data mining technique using the pretrained VGGish feature extractor and an anomaly detection network on long untrimmed audio files. The obtained datasets are used to trained a delicately designed super lightweight CNN for cry/non-cry classification. This CNN is then used as a feature extractor in an anomaly detection framework to achieve better cry detection performance.
\end{abstract}

\section{Introduction}

Baby monitoring is an important application of video surveillance, computer vision and machine learning. It helps take better care of babies and reduce the burden of care givers. Baby cry is a signal to communicate their needs including hunger, discomfort, or pain. It is used not only for the purpose of care giving, but also for purpose of disease diagnosis. 

The goal of is to detect the baby cry and localize its starting and end position in an audio signal. It is a challenging task since the baby cry sound may be mixed with different background noise in various environments, such as home and hospital.

In recent years, detection of baby cry has been studied using both traditional machine learning and deep learning algorithms. The traditional algorithm, typically SVM, works well on hand crafted acoustic features in both frequency domain and time domain. Typical examples include MFCCs and their varieties, pitch-related features, harmonic features, energy, zero crossing rate etc. For a review of these approaches, the readers are referred to \cite{CJi, Cohen2020, RTorres}. 

The deep learning algorithms mostly use CNN \cite{CJi, Cohen2020, RTorres, 18}, and some use other types of neural network \cite{9400952}. It has been shown in \cite {Cohen2020, RTorres} that the performance of CNN is much better than traditional machine learning. On the other hand, the complexity of the CNN may be high, preventing it be used in common embedded devices, like low-cost IP cameras or tablets. In this paper we will address this issue by designing a super lightweight CNN.   

Most, if not all, existing CNN methods use supervised learning. Therefore, frame-level annotation is needed. This annotation is very time consuming and prone to human mistakes. In the few datasets available online, some of the audio files \cite{giulbia, donateacry-corpus, esc50} are trimmed and annotated, some others \cite{audioset} are untrimmed with only audio-level annotations. In this work, we propose to use the weakly supervised anomaly detection \cite{UCF-Crime} to detect baby cry on audio signal. This weak supervision only requires weak annotation, i.e., if there is cry in the audio file without frame-level annotation, therefore makes the data annotation a lot easier. 

Based on this weakly supervised anomaly detection, and a well-known pretrained VGGish network \cite{vggish} for audio feature extraction, we design a data mining technique to obtain frame-level datasets for supervised CNN classification. We use this dataset to train a delicately designed super lightweight CNN, which runs very fast on embedded devices. This CNN, as the feature extractor in an anomaly detection framework, gives better performance than the CNN alone.       

The contribution of this paper is three-fold, 
\begin{itemize}
\item First we propose to use anomaly detection to detect baby cry in audio signals. To our knowledge, we are the first to use anomaly detection for the purpose of baby cry detection.
\item We design a data mining technique using the pre-trained VGGish \cite{vggish} feature extractor and an anomaly detector. The obtained dataset has similar performance to an annotated dataset on our lightweight CNN. 
\item We design a super lightweight CNN. This CNN makes our framework possible to run on embedded devices. 
\end{itemize}

\section{Related Work}

In this section we first review literature on baby cry detection. Then we review the anomaly detection approaches on video signals. We borrow this anomaly detection idea and extend it to baby cry detection on audio signals.   

\subsection{Baby Cry Detection}

The first step of baby cry detection is audio signal pre-processing. The main tasks are denoising and audio segmentation. The purpose of denoising is to filter-out noise in unwanted frequency band.  Audio segmentation is to use vocal activity detector (VAD) to remove silent duration. In this work we treat silent duration as non-cry segments and directly apply the cry detection on it.     

The signal processing features of audio signal can be categorized to cepstral domain, prosodic domain, time domain, image domain, and wavelet domain \cite{CJi}. The cepstral domain is the most widely used. It includes the Mel frequency cepstral coefficients (MFCCs), linear frequency cepstral coefficients (LFCCs), and the corresponding spectrogram. Please note that the spectrogram is 2D data, while the cepstral coefficients are a number of scalar data.   

Baby cry detection is essentially a binary classification task. A variety of classification techniques can be used, including 2-D CNN, 1-D CNN, SVM, KNN, multiple-layer perceptron (MLP). In previous work \cite{RTorres, 48, 25, 64, 16, 18, 87, 17}, these different features and classification methods are used on their private datasets. Since their datasets are private, it is impossible to conclude which is better. However, in their own comparison, a common recommendation is that the CNN outperforms the traditional machine learning methods. For a complete review, please refer to \cite{CJi, Cohen2020, RTorres}.  

As a super-set of baby cry detection, audio anomaly detection typically use unsupervised learning \cite{koizumi2020description}. The work \cite{9755147} presents a large audio dataset for anomaly detection including baby cry detection. However, even the term anomaly detection is used, in their detection algorithm, audio files are first cut into small segments then supervised learning is used on classification every segment. We are the first to explore weakly supervised learning for its significantly lower workload to prepare dataset.

\subsection{Anomaly Detection on Videos}

Weakly supervised anomaly detection only uses video-level annotations. These annotation only gives a binary label of abnormal or normal for a video. Sultani et al. \cite{UCF-Crime} propose the MIL framework using only video-level labels and introduce the large-scale anomaly detection dataset, UCF-Crime. This work inspires quite a few follow-up studies \cite{adgcn_cvpr19}, \cite{wsal_tip21}, \cite{UBI_Fight}, \cite{AAAI22}, \cite{CRFD}, \cite{IJCAI21}, \cite{MIST}, \cite{RTFM}, \cite{CRFD}. 
However, in the MIL-based methods, abnormal video labels are not easy to be used effectively. Typically, the classification score is used to tell if a snippet is abnormal or normal. This score is noisy in the positive bag, where a normal snippet can be mistakenly taken as the top abnormal event in an anomaly video. To deal with this problem, Zhong et al. \cite{adgcn_cvpr19} treat this problem as a binary classification under noisy label problem and use a graph convolution neural (GCN) network to clear the label noise. In RTFM \cite{RTFM}, a robust temporal feature magnitude (RTFM) is used to select the most reliable abnormal snippets from the abnormal videos and the normal videos.  They unify the representation learning and anomaly score learning by an temporal feature ranking loss, enabling better separation between normal and abnormal feature representations, improving the exploration of weak labels compared to previous MIL methods. More details will be given later. 

\section{Proposed Methods}

\begin{figure*}[tb]
    \centering
    \includegraphics[scale=0.675]{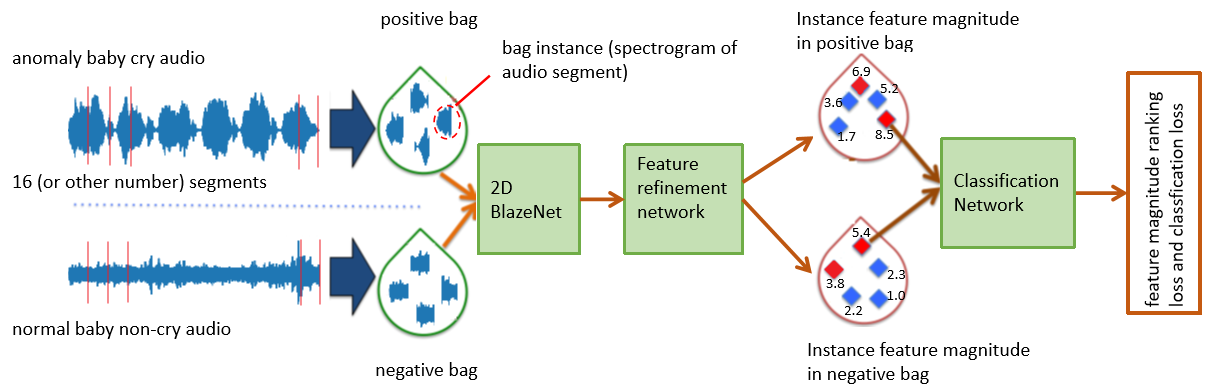}
    \caption{Anomaly detection of baby cry block diagram. The log-Mel-spectrom is not shown.}. 
    \label{fig1}
\end{figure*}

\subsection{Anomaly Detection in Audios}

The task of anomaly detection is to find and localize anomalous or abnormal events in videos. There are self-supervised methods trained only on normal datasets and weakly-supervised methods trained on both abnormal and normal datasets annotated with video or audio level labels. The weakly supervised anomaly detection in videos is first proposed in \cite{UCF-Crime}. It uses a multiple-instance learning (MIL) framework to find a segment in the positive (abnormal) or negative (normal) data sample whose classification scores are the maximum. Then the distance between the two segment scores are maximized for the best discriminability. 

In this work, we extend the anomaly detection from video signal to audio signal. In videos, since the frame is 2D, therefore a segment of frames is 3D. In audio signal, a segment is 1D audio signal and its spectrogram is 2D. So instead of a 3D CNN backbone, a 2D CNN backbone is needed.    

Let $V_a$ and $V_n$ represent the segments in the abnormal and normal audio. The MIL expects to have the following objective function,  

\begin{equation}
    \max\limits_{i \in B_a} f(V_a^i) >  \max\limits_{i \in B_n} f(V_n^i) 
    \label{eq1}
\end{equation}
where $B_a$ and $B_n$ are the bags of segments in the abnormal and normal audio, $f$ is the predicted anomaly score in range of 0 and 1. This function $max$ is taken over all instances in a bag. It is used because the segment level annotation is not available. It is expected that in the positive bag, the highest-scored instance is a true abnormal segment. The highest-scored instance in the negative bag is the one most similar to the positive bag, but is actually a negative instance. This makes the negative instance a hard one and therefore benefits the discriminability in the model training. To push the positive instance and negative instance further apart, the MIL ranking loss is defined as,  

\begin{equation}
    l(B_a,B_n) = \max(0, 1-\max\limits_{i \in B_a} f(V_a^i) + \max\limits_{i \in B_n} f(V_n^i))
    \label{eq2}
\end{equation}
It is worth noting that this loss function looks similar to the contrastive loss function which is used to separate two or more classes as farther as possible. Two regularization terms, the smoothness term and the sparsity term are added onto it. So the overall loss function is \cite{UCF-Crime}, 

\begin{align}
    l = & \max(0, 1-\max\limits_{i \in B_a} f(V_a^i) + \max\limits_{i \in B_n} f(V_n^i))\nonumber \\
        & +\lambda_1 \sum_i (f(V_a^{i+1})-f(V_a^i))^2+\lambda_2 \sum_i (f(V_a^i))^2
    \label{eq3}
\end{align}

It is expected in Eq. \ref{eq1} that abnormal segments have higher score than normal segments. However, this is not always true. A few methods \cite{wsal_tip21}, \cite{AAAI22}, \cite{CRFD}, \cite{IJCAI21}, \cite{MIST}, \cite{RTFM}, \cite{CRFD} have been studied how to improve the score quality so that the correct abnormal segment is chosen in the abnormal bag. In the work RTFM \cite{RTFM}, a different approach is used. Instead of using the classification score as the criterion to choose the abnormal segment, the authors propose to use a feature magnitude, which they believe has better discriminability between abnormal and normal instances. Furthermore, they propose to use multiple instances whose feature amplitude are the largest and call them the top-k instances. In their approach, the MIL ranking loss is defined on the feature magnitude, 

\begin{equation}
    l_F(B_a,B_n) = \max(0, m-d(m(V_a^{top-k}), m(V_n^{top-k}))
    \label{eq4}
\end{equation}
where $X_a^{top-k}$ and $X_n^{top-k}$ are the top-k segments whose feature magnitudes are the largest k instances out of the abnormal and normal bag, $f$ is the feature magnitude function, $m$ is a pre-defined margin, and $d$ is the defined distance function between the two sets of top-k features. In their implementation, this function is simply the square of mean of the top-k feature magnitude. 

The standard cross-entropy loss is used as the classification loss. However, it is applied on the top-k segments whose feature magnitude are the largest. If $k>1$, the scores are averaged before feeding into the cross-entropy loss function,

\begin{equation}
    l_S(B_a,B_n) = -ylog(f(X^{top-k})-(1-y)log(1-f(X^{top-k}))
    \label{eq5}
\end{equation}
The same smoothness term and sparsity term are also used, so the overall loss function is,

\begin{align}
    l(B_a,B_n) =  & l_S+\alpha l_F \nonumber \\
     &+\lambda_1 \sum_i (f(V_a^{i+1})-f(V_a^i))^2+\lambda_2 \sum_i (f(V_a^i))^2
    \label{eq6}
\end{align}

where $\alpha$,$\lambda_1$ and $\lambda_2$ are pre-defined weight factors. 

In addition, a multi-scale (dilated convolution) non-local aggregation (MSNL) block is used \cite{RTFM} on the feature extracted from the pre-trained CNN backbone. This block is also important for the feature magnitude training. Without this block the feature is fixed and cannot be learned. The MSNL is used in \cite{RTFM}, but other simpler network, e.g., a few full-connection (FC) layers may also work. 

\subsection{Proposed Anomaly Detection Framework}

The overall block diagram of our proposed network is illustrated in Fig. \ref{fig1}. The style of the figure is borrowed from \cite{UCF-Crime}. A framework similar to RTFM \cite{RTFM} is used, with all necessary modifications for anomaly detection of baby cry in audios. 

The abnormal or normal audio signal is first divided to a certain number of equal-length segments. We use 16 in the figure as an example. Every segment is called an instance in the positive or negative bag of instances. All instance pass through a pre-trained CNN backbone, and CNN features are extracted. In Figure 1, the CNN backbone is called BlazeNet - our delicately designed super lightweight network, whose detail will be given later.  This CNN feature is fed into a feature refinement network and a second CNN feature is extracted. The features of the positive instance bag form the positive feature bag, same is for the negative feature bag. The top-k instances whose feature magnitudes are the largest among this bag are selected. The classification network is typically two or three FC layers. Through the back propagation of the loss function in Eq. \ref{eq5}, the feature magnitude and the classification are both learned at the same time.

In implementation, when the audio file is very short and the audio signal is divided into 16 segments, every segment may not be long enough for a frame. So CNN feature is extracted for every frame, then linear interpolation is used to generate features for 16 segments. 

\subsection{BlazeNet}

Our goal of this study is to design a baby cry detection framework that can work efficiently on embedded devices. So the CNN backbone network must be super lightweight, and at the same time, achieve good performance. We try the popular MobileNet, ShuffleNet, SqueezeNet, and find that they are still too large, no need to mention the popular VGGish-Net widely used in the audio recognition. 
We take the backbone from the BlazeFace in \cite {BlazeFace}. We make changes so that the input size is 64x64 and the output feature size is 224. We use 16 BlazeBlocks, where the 11th BlazeBlock output is classified by first classifier FC1(88,2), and the 16th BlazeBlock output is classified by a second classifier FC2(96,6). The outputs of these two classifiers are flattened and concatenated then classified by the final FC(224,2) classifier. The details of our BlazeNet are listed in Table 1. The total number of parameters of this model is 89,680 in PyTorch. 

\begin{table}
  \centering
  \begin{tabular}{|c|c|}
    \hline
    BlazeBlock  &  Others\\
    \hline
     - & Conv2D(3,24,5,2) \\
    BlazeBlock-1 (24,24,3,1) & -\\
    BlazeBlock-2 (24,28,3,1) & -\\
    BlazeBlock-3 (28,32,3,2) & -\\
    BlazeBlock-4 (32,36,3,1) & -\\
    BlazeBlock-5 (36,42,3,1) & -\\
    BlazeBlock-6 (42,48,3,2) & -\\
    BlazeBlock-7 (48,56,3,1) & -\\
    BlazeBlock-8 (56,64,3,1) & -\\
    BlazeBlock-9 (64,72,3,1) & -\\
    BlazeBlock-10 (72,80,3,1) & -\\
    BlazeBlock-11 (80,88,3,1) & FC1(88,2)\\
    BlazeBlock-12 (88,96,3,2) & -\\
    BlazeBlock-13 (96,96,3,1) & -\\    
    BlazeBlock-14 (96,96,3,1) & -\\
    BlazeBlock-15 (96,96,3,1) & -\\
    BlazeBlock-16 (96,96,3,1) & FC2(96,6)\\
    -              &     cat(FC1,FC2) \\
    - &                   FC(224,2) \\
    \hline

  \end{tabular}
  \caption{Layers of out BlazeNet. The BlazeNet parameters are (number of input channels, number of output channels, kernel size, stride). The sequence of layers is from top to bottom, from left to right.} 
  \label{Table-1}
\end{table}

\begin{figure}[tb]
    \centering
    \includegraphics[scale=0.875]{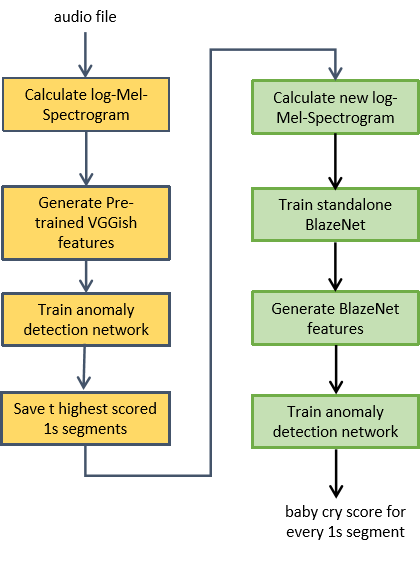}
    \caption{pipeline of data mining and anomaly detection for baby cry detection}. 
    \label{fig2}
\end{figure}

\subsection{Data Mining Datasets for BlazeNet}

The CNN backbone in Figure 1 is pre-trained and fixed when the anomaly detection network (feature refinement network and classification network) is trained. If this CNN and the anomaly detection network are trained together end-to-end, the GPU can be easily overflowed. For this reason, all previous anomaly detection methods in videos use pre-trained 3D CNN. 

So we need to pre-train the CNN backbone before the whole anomaly detector framework can work to detect baby cry in audios. To do so, we need some training data (herein including validation and test dataset without confusion). One way to do so is to prepare some audio data manually, which we do in this work. More details will be given later. However, manually annotating audio data is very time consuming and is prone to human mistakes. We propose a second way to mine training data from weakly-annotated dataset and a different pre-trained CNN backbone.  

In \cite{audioset}, the authors publish a large audio dataset called AudioSet. At the same time they publish a pre-trained VGGish backbone \cite{vggish} for CNN feature extraction. With their default settings, the log-Mel-spectrogram is used on a 0.96-s frame. The output CNN feature is 128-D. We use this VGGish network to exact CNN features for audio files, then apply the anomaly detection framework onto it. So to make it clear, the 2D BlazeNet in Figure 1 is replaced with the pre-trained VGGish network. After the anomaly detection network (feature refinement network and classification network) is trained, the framework is set to inference mode, and all training, validation, and test datasets are processed. Please note that, in the inference mode, the audio files are not divided into 16 segments. In stead, the audio signal in the form of 1s-frame is used.  Then the top $t$ 1s segments whose classification scores are the largest are saved. In this way we obtain the new training, validation, and test dataset to train the 2D BlazeNet. 

The pipeline of the data mining and anomaly detection frame work for baby cry is shown in Figure 2. The blocks on the left perform the data mining, and the blocks on the right perform the anomaly detection of baby cry detection. Please note only the training procedure of anomaly detection is plotted, and the testing procedure can be derived accordingly.

\begin{table*}
  \centering
  \begin{tabular}{|c|c|c|c|c|}
    \hline
    Source  &  length  & annotation & cleaned number & background\\
    \hline
    \cite{giulbia} &  5s & Baby Cry, other 3 & 108, 324  & clean \\
    \cite{donateacry-corpus} &  7s & Baby Cry only & 482 & clean \\
    ESC-50 \cite{esc50} & 5s & Baby Cry, many others &40, 1960 & clean \\ 
    AudioSet \cite{audioset} & Untrimmed & Baby Cry, many others & 1364, a lot & noisy\\
    \hline
  \end{tabular}
  \caption{Dataset sources of baby cry we find online.} 
  \label{Table-2}
\end{table*}

\section{Experiments}

\subsection{Datasets}

Even though there are quite some publications on baby cry detection, none of the used dataset are publicly available. We search online and organize the datasets from the sources listed in Table 2. Please note that the background of all these datasets are clean except for the AudioSet \cite{audioset}, whose background is very noisy. We clean the datasets and filter out the ambiguous cases. The numbers of cleaned samples are listed in the table. We will publish the dataset we organize soon.

In total we have 2624 baby cry audios and tremendous other audios. We use all other audios in \cite{giulbia,donateacry-corpus,esc50}, then collect randomly other audios in \cite{audioset}. The total number of other non-cry audios is about the same as the number of baby cry audio. 

For the training of BlazeNet, we have two audio frame lengths, 5s and 1s. If the length of an audio is longer than two times the frame length, them more than one frames can be cut from an audio file. After the cutting we manually check if there is really a segment of baby cry in the audio file. The total audio frames are randomly divided into training, validation and test datasets with ratio 8:1:1. 

For the anomaly detection, we use two audio lengths, one is 5s, the other is the original audio file length. When the length is 5s, it is the same dataset as above. In this case, the frame length is only 1s. When the length is the original audio file length, the frame length is also 1s.  

Until very recently, we find that a new baby cry dataset is released in \cite{austin}. However, the link to where the dataset is saved is broken. So it is not really publicly available. 

\subsection{Implementation Details}

For the BlazeNet as a standalone CNN classification network on baby cry detection, we implement it in PyTorch. The SGD is used as the optimizer with starting learning rate 0.001 and momentum 0.9. The training runs 60 epochs. The learning rate is decayed by factor 0.1 every 20 epochs. A batch size 32 is used.  

For anomaly detection, we use the RTMF codebase \cite{RTFM} in PyTorch. Every audio file is divided to 5 segments when the input audio file is 5s long. It is divided to 10 segments when the original audio files are used. The top-k is set to 2. Two dataset iterators, one for the abnormal data and the other for the normal data, are used. This way, the pairing of abnormal and normal data is random, even when the numbers of abnormal and normal samples are different. An initial training rate of 1E-3 is used, and the training runs 20000 steps (we do not use epochs because the abnormal data loader and the normal data loader are iterating). A batch size 128 is used. 

For the VGGish input, the default log Mel spectrogram parameters are used. Specifically, sampling rate = 16K Hz, number of frames in batch = 96, number of Mel bands = 64, FFT window length = 0.025s, FFT hop length = 0.01s, min Mel frequency = 125 Hz, max Mel frequency = 7500 Hz, log offset=0.01, example hop seconds = 0.96. The only change we make is example window seconds = 1s so that we have a 96x64 log Mel spectrogram output for every 1s of audio. 

For the BlazeNet input, we use different log Mel spectrogram parameters and we use the Librosa library. When the example window seconds = 1s, sampling rate = 8K Hz (this is the audio signal sampling rate on most IP cameras), number of Mel bands = 64, FFT window length = 0.064s, FFT hop length = 0.01475, min Mel frequency = 0 Hz, max Mel frequency = 8000 Hz. Other default parameters are used. Please note that we use a FFT hop length such that the spectrogram out size is 64x64, which is required by the BlazeNet. Resizing the Mel spectrogram array is not recommended since it cause performance loss. When the example window seconds = 5s, the FFT window length and FFT hop length are adjusted accordingly so the spectrogram out size is 64x64. 

Please note that, in all our performance evaluations, the data unit the 1s audio segment. In practice, we give detection result for every 1s audio signal input.   

\begin{table}[tb]
\begin{center}
  \begin{tabular}{|c|c|c|c|}
    \hline
    Dataset  &  Val Acc & Test Acc   \\
    \hline
    5s audios & 0.9447 & 0.9312\\
    untrimmed audios & 0.9370 & 0.9223\\
  \hline
  \end{tabular}
  \caption{Performance of anomaly detection of baby cry using VGGish features. Accuracy is measured at default classification threshold = 0.5.}
  \label{table-3}
  \end{center}
  \end{table}

\begin{table*}[tb]
\begin{center}
  \begin{tabular}{|l|c|c|c|c|l|}
    \hline
    Dataset  & Val Acc & Test Acc   \\
    \hline
    all 1s segments from 5s audios & 0.8868 & 0.8820 \\
    random 2 1s-segments from 5s audios & 0.8784 & 0.8654\\
    \hline
    mined top-2 1s-segments from 5s audios  & 0.8609 & 0.8542 \\
    mined positive top-2 1s-segments from 5s audios  & 0.8748 & 0.8562 \\
    mined negative top-2 1s-segments from 5s audios  & 0.8748 & 0.8622 \\
    \hline
    mined top-2 1s-segments from untrimmed audios  & 0.8142 & 0.8292 \\
    \hline

  \hline
  \end{tabular}
  \caption{Performance of BlazeNet as a standalone baby cry classifier. Accuracy is measured at default classification threshold = 0.5.}
  \label{table-4}
  \end{center}
  \end{table*}

\subsection{Data Mining using VGGish}

We first test how VGGish \cite{vggish} features work in the anomaly detection network. The performance must be good for the data mining to work well. From the standpoint of anomaly detection, positive instances must have larger score than the negative instances (see Eq. 1). In other words, the largest scored instances in the abnormal bag must be truly positive instances. 

We do this test on both the trimmed 5s audio files and the untrimmed audio files as dataset.  VGGish features are extracted for 1s segments sequentially without overlap in every audio file. These features and their audio labels are used in training and testing the anomaly detection network. 

The experiment results are listed in Table 3. We observe that the validation and test accuracy results are all higher than 0.90. We believe this good performance will make the data mining method work well, which will be verified later with the performance of the standalone BlazeNet classification. 

\subsection{BlazeNet Classification Results}

We first train the BlazeNet on trimmed 1s audios. The trimmed 5s audios are prepared manually. To get 1s audio dataset, every 5s audio is cut into 5 segments of 1s-long audios without overlap. When the 1s audios are used in training, there are two modes. In the first mode, all 5 segments out of every 5s long audio file are used. In the second mode, only 2 randomly selected segments are used. This is for a fair comparison with the data mining method, where only the top-2 segments are saved as training dataset.  

Please note that in all these experiments, only the training dataset changes, while the validation and test datasets stay the same. In the last experiment, long untrimmed audios are added to the training dataset. We argue that always using short 1s audio files as validation and test datasets is reasonable because in practical applications, a decision per 1s audio signal is preferred to avoid long latency.  

The experiment results are listed in Table 4, where all the accuracy results are taken at default threshold = 0.5. The accuracy result of using all 1s segments is the best, and the one using random 2 segments is a little bit worse. This is probably because the selected segments do not cover as many cases as using all 1s segments.  

Considering the 5s audios are well annotated, any 1s segment should be good to be used. When top-2 1s segments are mined from 5s audios, we expect to have same performance as the random 2 1s segments, however the results are not so. The accuracy is worse than using the annotated data by 1\%. We test further two cases, one using only mined positive data, the other using only mined negative data. The results show that hard negative samples with highest scores in the mined negative data are preferred, while the positive samples with highest scores are not preferred.   

The performance of the mined data from the long untrimmed audios is even worse. This is understandable since some negative samples may be chosen as positive samples. However, as a backbone for an anomaly detection framework, the backbone does not need to be perfect. As in anomaly detection in videos, an pre-trained 3D CNN is used without training on the anomaly detection dataset \cite{UCF-Crime, RTFM}. Experiment result will be shown in next subsection.

\begin{table}[tb]
\begin{center}
  \begin{tabular}{|c|c|c|c|}
    \hline
    Dataset  &Backbone & Val Acc & Test Acc   \\
    \hline
    5s audios & Table 4 Line-1 &  0.9457 & 0.9302\\
    untrim audios & Table 4 Line-1 &  0.9141 & 0.9020\\
    untrim audios & Table 4 Line-6 &  0.8652 & 0.8473\\
  \hline
  \end{tabular}
  \caption{Performance of anomaly detection of baby cry using BlazeNet features. Two trained BlazeNet backbones from Table 4 are used. Accuracy is measured at default classification threshold = 0.5.}
  \label{table-5}
  \end{center}
  \end{table}

\begin{table}[tb]
\begin{center}
  \begin{tabular}{|c|c|c|c|}
    \hline
    Dataset  & Method & F1-max  &  Test Acc  \\
    \hline
    5s audios & BlazeNet & 0.8873 & 0.8836\\
    \hline
    5s audios & Anomaly & 0.9241 & 0.9305\\
    untrim audios & Anomaly& 0.9306 & 0.9360\\
  \hline
  \end{tabular}
  \caption{Performance comparison of standalone BlazeNet and anomaly detection with BlazeNet backbone from Table-4 Line-1. Test accuracy is at the threshold where F1-max is achieved.}
  \label{table-6}
  \end{center}
  \end{table}

\subsection{Anomaly Detection Results}

In this subsection we test the anomaly detection framework shown in Figure 1 with the fixed BlazeNet, which has been trained in the previous subsection. 

In the first experiment, we use this method on the manually annotated 5s audio files. The goal is to find the detection results on every 1s audio segments and measure the accuracy. In this experiment, since there are only 5 1s segments in a 5s audio file, the number of segments in anomaly detection is set to 5, and the top-k is set to 2. 

In the second experiment, long untrimmed audio files are used as training dataset, while 5s audio files are used as validation and test datasets. After looking at the distribution of the audio file length, we set the number of segments in anomaly detection to 10. The the top-k is still set to 2. 

The experiment results are listed in Table 5. We observe that the performance is very close to that using VGGish feature in Table 3, while the complexity of the BlazeNet is a lot lower than that of the VGGish network. For BlazeNet trained at Table-4 Line-6, the performance is worse than the one trained at Table-4 Line-1. So the quality of the BlazeNet feature does matter in the anomaly detection performance. So manually annotating some dataset for the BlazeNet is preferred.    

\subsection{Discussion: Anomaly Detection vs. Classification} 

Our goal is to find a solution for baby detection on embedded devices, so we ignore any results directly using VGGish network in inference mode. When comparing the results of BlazeNet in Table 4 and Table 5, we see that the performances of anomaly detection is already more than 2\% better than of the standalone BlazeNet. 

Furthermore, we note that we only use the accuracy at default threshold = 0.5 in all these experiments (We use this threshold because it is dominantly used in training of a binary classifier). So we do some analysis in terms of max F1 score and the ROC curve. 

We collect prediction scores of all 1s audio segments in the test dataset, then calculate the max F1 score and the corresponding threshold. Finally we calculate the test accuracy at this threshold. The results are listed in Table 6. It is observed that for the BlazeNet, since it is a binary classification, the threshold to achieve the max F1 is near 0.5, and the test accuracy at this threshold is almost identical to the one in Table 4. While for anomaly detection, for its nature of MIL ranking loss, the threshold to achieve the max F1 is pushed to the 1.0 side.   

The ROC curves of the same three cases in Table 6 are plotted in Figure 3. The gain from the anomaly detection is obvious, so is the ROC AUC.  

\begin{figure}[tb]
    \centering
    \includegraphics[scale=0.6]{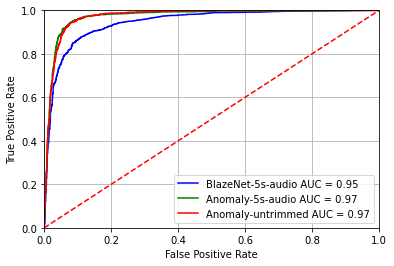}
    \caption{ROC curves of standalone BlazeNet and the anomaly detection.} 
    \label{fig3}
\end{figure}
\section{Conclusion}

In this paper, we study the baby cry detection in audios. We design a super lightweight BlazeNet for baby cry/non-cry classification. On top of that, we propose to use anomaly detection to do the same task. Experiment results show that the anomaly detection can achieve better performance than the standalone BlazeNet with a little bit extra complexity. 

We further propose to use the anomaly detection framework with a pre-trained VGGish backbone to mine training data for the BlazeNet. Even though the BlazeNet trained with this data is not as well as the one using manually annotated data, using this BlazeNet in an anomaly detection framework still work relatively well. This can save the time consuming work to annotate audio data at the frame level. Instead, audio file level weak label can be used. 

{\small
\bibliographystyle{ieee}
\bibliography{egbib}
}

\end{document}